%% file: main.tex
\documentclass[runningheads]{llncs}
\usepackage{graphicx}
\newcommand*\samethanks[1][\value{footnote}]{\footnotemark[#1]}
\usepackage[table]{xcolor}
\usepackage{chngpage}
\usepackage{float}
\usepackage{etoolbox}
\usepackage{amsmath}
\usepackage{rotating}
\usepackage{graphicx}

\apptocmd{\sloppy}{\hbadness 10000\relax}{}{}

\begin{document}

\title{FedPIDAvg: A PID controller inspired aggregation method for Federated Learning}

\input{authors.tex}

\maketitle
\setcounter{footnote}{0}
\begin{abstract}

This paper presents FedPIDAvg, the winning submission to the Federated Tumor Segmentation Challenge 2022 (FETS22). Inspired by FedCostWAvg, our winning contribution to FETS21, we contribute an improved aggregation strategy for federated and collaborative learning. FedCostWAvg is a weighted averaging method that not only considers the number of training samples of each cluster but also the size of the drop of the respective cost function in the last federated round. This can be interpreted as the derivative part of a PID controller (proportional–integral–derivative controller). In FedPIDAvg, we further add the missing integral term. Another key challenge was the vastly varying size of data samples per center. We addressed this by modeling the data center sizes as following a Poisson distribution and choosing the training iterations per center accordingly. Our method outperformed all other submissions.

\vspace{0.6cm}
\end{abstract}
\newpage

\input{intro.tex}
\input{method.tex}

\input{results.tex}

\input{acknowledgment}

{ \bibliographystyle{splncs}
\bibliography{mybibliography}
}

\end{document}

%% file: authors.tex
\author{Leon M\"achler\inst{6} \and
Ivan Ezhov\inst{1,3} \and
Florian Kofler\inst{1,2,3} \and
Suprosanna Shit\inst{1,3} \and
Johannes C. Paetzold\inst{1,3,5} \and
Timo Loehr \inst{1,3} \and
Claus Zimmer\inst{2} \and
Benedikt Wiestler\inst{2} \and
Bjoern H. Menze\inst{1,3,4}\samethanks}



\author{Leon M\"achler\inst{4} \and
Ivan Ezhov\inst{1,2} \and
Suprosanna Shit\inst{1,2} \and
Johannes C. Paetzold\inst{1,2,3}}

\authorrunning{M\"achler et al.}
\titlerunning{FedPIDAvg}

%
\institute{Department of Informatics, Technical University Munich, Germany\and 
TranslaTUM - Central Institute for Translational Cancer Research, Technical University of Munich, Germany\and
ITERM Institute Helmholtz Zentrum Muenchen, Neuherberg, Germany \and
Département d'Informatique de l'ENS (DI ENS), École Normale Supérieure, PSL University, Paris, France \\
\email{leon-philipp.machler@ens.fr}\\
}


%

%% file: intro.tex
\section{Introduction}

Federated learning is a highly promising approach for privacy, and confidential learning across multiple data locations \cite{yang2019federated}. A vast set of applications exist, ranging from power grids to medicine \cite{li2020review}. Evidently, such approaches are of paramount importance for medical images, because patient information can be highly sensitive \cite{rieke2020future} and the distribution of medical expertise, as well as the prevalence of certain diseases, is extremely uneven. More practically, medical imaging data is extremely large in size, making the frequent transfer of data from a local clinic to a central server location very expensive \cite{rieke2020future}. Privacy and safety of patient data is even more emphasized when we consider the large illegal leaks of private medical records to the \emph{dark web}\cite{meddatahack}.

\begin{figure}[ht!]
    \label{fl}
    \centering
    \includegraphics[width=0.9\textwidth]{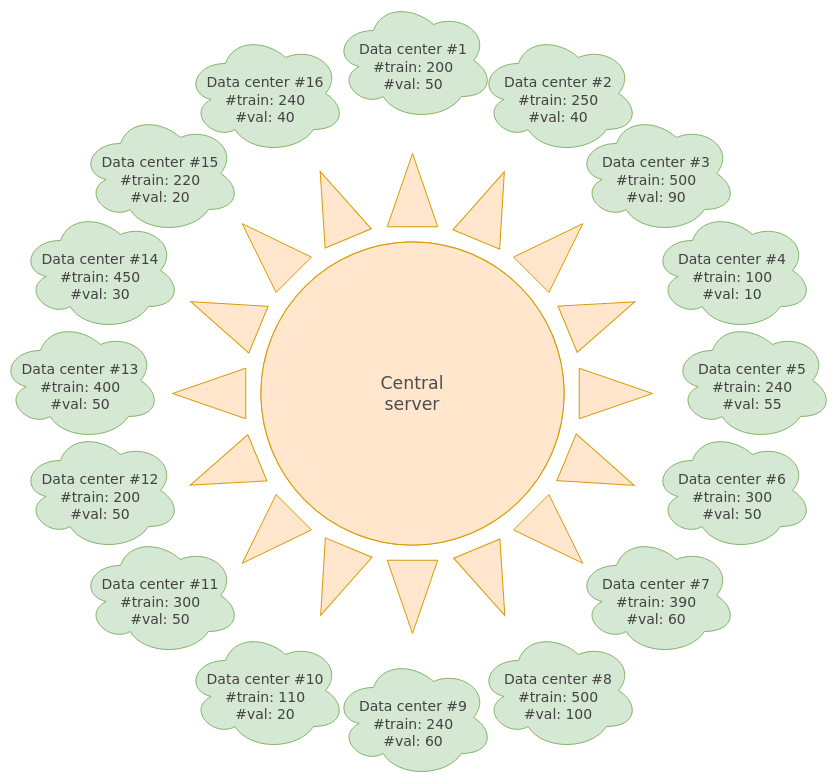}
    \caption{Schematic illustration of the federated learning concept. One can see how multiple data centers make up one big federation. The training data is stored exclusively at the local centers, where the same model is trained locally for a defined task. E.g., brain tumor segmentation as in our case. In the aggregation step, the model weights are sent and collected at a central server location. Here, the model aggregation is performed and later broadcasted back to the local centers. This procedure is repeated until convergence or another stopping criteria is reached.}
    \label{fig2}
\end{figure}

\subsection{FETS challenge}
The FETS challenge \cite{bakas2017advancing,pati2021federated,reina2021openfl,sheller2020federated,baid2021rsna} is an initiative trying to address the main research question of federated learning: optimal aggregation of network weights coming from various data centers. In this paper, we try to address this issue by proposing a PID and classical statistics-inspired solution.

\section{Prior work}
\subsection{Federated Averaging (FedAvg)}

The traditional federated averaging (FedAvg) approach \cite{mcmahan2017communication} employs an averaging strategy on the local model weights to update the global model, weighted by the training data set sizes of the local models. A model $M_{i+1}$ is  updated as follows: 

\begin{equation}
    M_{i+1} = \frac{1}{S} \sum_{j=1}^n s_j M_{i}^j. 
\end{equation}\\

Here, $s_j$ is the number of samples that model $M^j$ was trained on in round $i$ and $S = \sum_j s_j$. The definition is adapted from \cite{machler2021fedcostwavg}.

\subsection{Federated Cost Weighted Averaging (FedCostWAvg)}

Last year, we proposed a new weighting strategy, which won the FETS21 challenge. It not only weighs by data center sizes but also by the amount by which the cost function decreased during the last step \cite{machler2021fedcostwavg}. We termed this method FedCostWAvg, where the new model $M_{i+1}$ is calculated in the following manner: 

\begin{equation}
    M_{i+1} = \sum_{j=1}^n (\alpha \frac{s_j}{S} + (1 - \alpha) \frac{k_j}{K}) M_{i}^j.
\end{equation}
with:
\begin{equation}
     k_j = \frac{c(M_{i-1}^j)}{c(M_{i}^j)}, K = \sum_j k_j.
\end{equation}\\

Here, $c(M_i^j)$ is the cost of the model $j$ at time-step $i$, which is calculated from the local cost function \cite{machler2021fedcostwavg}.  Moreover, $\alpha$ ranges between $[0,1]$ and is chosen to balance the impact of the cost improvements and the data set sizes. Last year, we won the challenge with an alpha value of \emph{$\alpha = 0.5$}. We discussed that this weighting strategy adjusted for the training dataset size and also for local improvements in the last training round. 

%% file: method.tex
\section{Methodology}

In the following chapter, we will first introduce and formalize our novel averaging concept named \textit{FedPIDAvg}. Next, we will quickly describe the neural network architecture for brain tumor segmentation that was given by the challenge organizers and finally discuss our strategy regarding when to train and aggregate from which specific centers, depending on their training samples modeled using a simple Poisson distribution. 

\subsection{Federated PID Weighted Averaging (FedPIDAvg)}

As was already mentioned in \cite{machler2021fedcostwavg} David Nacacche offered the observation that the idea of FedCostAvg is similar to that of a PID controller. Only the integral term is missing. The methodology of our new averaging method is novel in two ways: the PID-inspired added integral term, and a different way to calculate the differential term. Our method calculates the new model $M_{i+1}$  in the following manner: 

    \begin{equation}
    M_{i+1} = \sum_{j=1}^n (\alpha \frac{s_j}{S} + \beta \frac{k_j}{K} + \gamma \frac{m_j}{I} ) M_{i}^j.
\end{equation}
with:
\begin{equation}
     k_j = c(M_{i-1}^j) - c(M_{i}^j), K = \sum_j k_j.
\end{equation}
and:
\begin{equation}
     m_j = \sum_{l=0}^5 c(M_{i-l}) , I = \sum_j m_j.
\end{equation}\\
\begin{equation}
    \alpha + \beta + \gamma = 1
\end{equation}\\

Note that this time we use the absolute difference between the last cost and the new cost and no longer the ratio. The new strategy is still a weighted averaging strategy, where the weights are themselves weighted averages of three factors: the drop of the cost in the last round, the sum over the cost in the last rounds and the size of the training data set. These three factors are weighted by $\alpha, \beta$ and $\gamma$. Their choice needs to be optimized based on the use case, we chose $0.45, 0.45, 0.1$ although we did not have the resources to cross validate them.


\subsection{U-Net for Brain Tumor Segmentation}

As a segmentation architecture, we were given by the organizers the 3D-Unet, a vastly successful neural network architecture in medical image analysis \cite{ronneberger2015u}. No modifications to the architecture were allowed in the challenge, we quickly depict it in Figure \ref{fig2} for completeness. U-Nets constitute the state of the art for a vast set of applications, for example, brain tumor segmentation \cite{menze2014multimodal,bakas2017advancing}, vessel segmentation \cite{vessap,cldice} and many more.


\begin{sidewaysfigure}[t!]
    \centering
    \includegraphics[width=\textwidth]{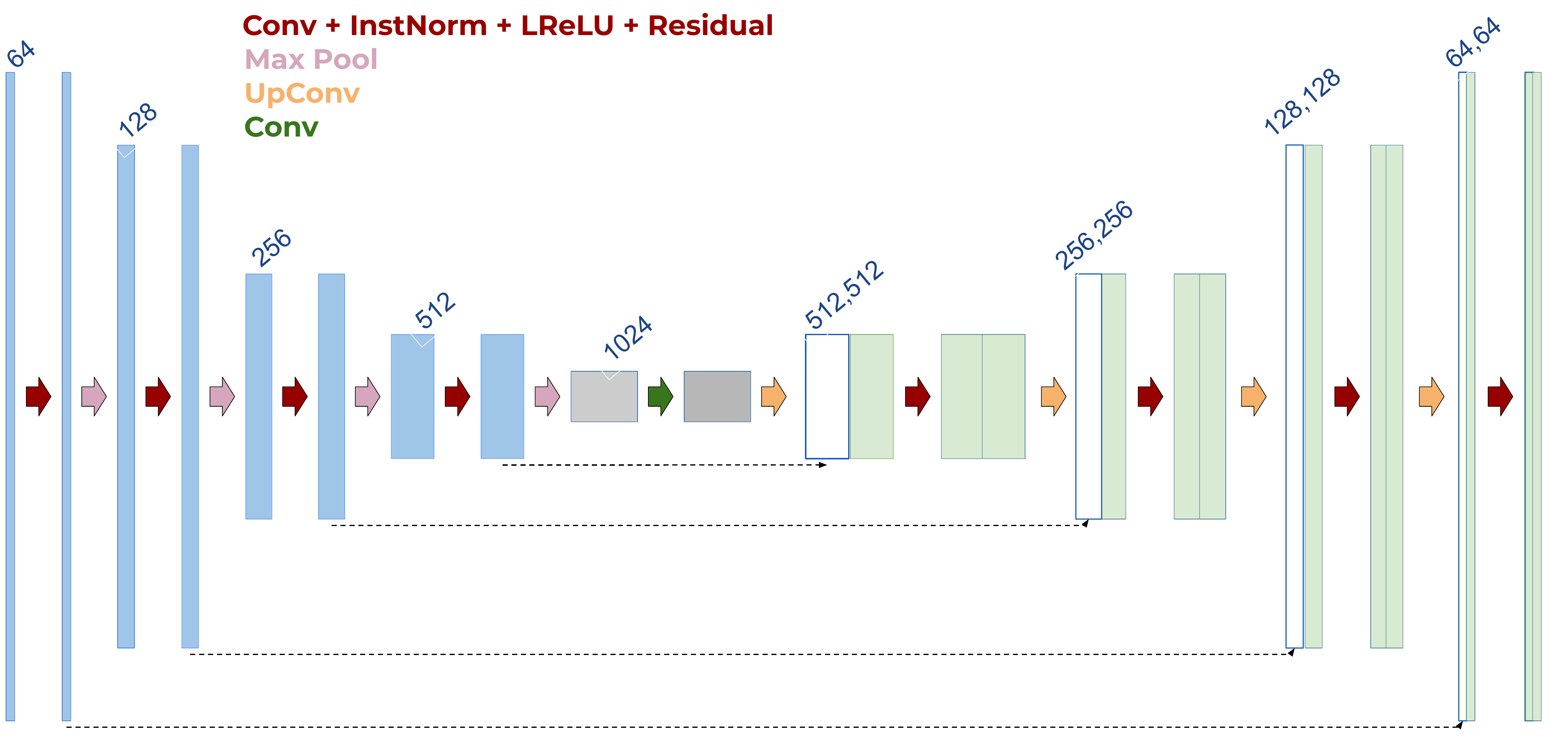}
    \caption{The common 3D U-net architecture, which is used in many medical imaging tasks. It was provided as such by the FETS challenge, modifications were not allowed \cite{ronneberger2015u,pati2021federated}.}
    \label{unet}
\end{sidewaysfigure}

\subsection{Poisson-distribution modeling of the data samples per center} In order to optimize the training speed over several federated rounds, it was possible to only select a subset of data centers for each federated round. In our last year's submission, we simply selected all centers every time. Another part of our submission this year is a novel way to select participating data centers at each federated round. To do so, we resorted to classical statistics means, namely, under the assumption that the dataset sizes follow a Poisson distribution:
\begin{equation}
\begin{aligned}
p(x;\lambda) = \frac{e^{-\lambda}\lambda^{x}} {x!}  \\ \\ \mbox{    with     } 
x = 0, 1, 2, \cdots
\end{aligned}
\end{equation} \\
we made the natural choice of dropping out outliers in most rounds, where outliers were defined as having $x>2\lambda$.



%% file: results.tex
\section{Results}
The methods were evaluated on the data of the Fets challenge. It is desribed as: "FeTS borrows its data from the BraTS Continuous Evaluation, but additionally providing a data partitioning according to the acquisition origin for the training data. Ample multi-institutional, routine clinically-acquired, pre-operative baseline, multi-parametric Magnetic Resonance Imaging (mpMRI) scans of radiographically appearing glioblastoma (GBM) are provided as the training and validation data for the FeTS 2022 challenge. Specifically, the datasets used in the FeTS 2022 challenge are the subset of GBM cases from the BraTS Continuous Evaluation. Ground truth reference annotations are created and approved by expert board-certified neuroradiologists for every subject included in the training, validation, and testing datasets to quantitatively evaluate the performance of the participating algorithms." Amongst all submitted methods, FedPIDAvg performed best and won the challenge.

In tables \ref{tab:f21.1} and \ref{tab:f21.2} we give last years results of FedCostWAvg in the last challenge and in tables \ref{tab:f22.1} and \ref{tab:f22.2} the results of FedPIDAvg in FETS22. Note that the data was different this year as well as the limits on available federated rounds.

\begin{table}[H]
\begin{adjustwidth}{0in}{0in}
    \centering
    \setlength\tabcolsep{3pt} 
    \begin{tabular}{|l|c|c|c|c|c|c|c|c|c|c|c|c|c|c|}
        \hline
        
        Label & Dice WT & Dice ET & Dice TC & Sens. WT & Sens. ET & Sens. TC   \\ \hline
        Mean & 0,8248 & 0,7476 & 0,7932 & 0,8957 & 0,8246 & 0,8269  \\ \hline 
        StdDev & 0,1849 & 0,2444 & 0,2643 & 0,1738 & 0,2598 & 0,2721  \\ \hline
        Median & 0,8936 & 0,8259 & 0,9014 & 0,948 & 0,9258 & 0,9422  \\ \hline
        25th quantile & 0,8116 & 0,7086 & 0,8046 & 0,9027 & 0,7975 & 0,8258 \\ \hline
        75th quantile & 0,9222 & 0,8909 & 0,942 & 0,9787 & 0,9772 & 0,9785  \\ \hline
    \end{tabular}
    \vspace{2pt}
    \caption{\label{tab:f21.1} Final performance of FedCostWAvg in the 2021 FETS Challenge, DICE and Sensitivity}
\end{adjustwidth}
\end{table}

\begin{table}[H]
\begin{adjustwidth}{0in}{0in}
    \centering
    \setlength\tabcolsep{3pt} 
    \begin{tabular}{|l|c|c|c|c|c|c|c|c|c|c|c|c|c|c|}
        \hline
        Label & Spec WT & Spec ET & Spec TC & H95 WT & H95 ET & H95 TC & Comm. Cost  \\ \hline
        Mean & 0,9981 & 0,9994 & 0,9994 & 11,618 & 27,2745 & 28,4825 & 0,723 \\ \hline
        StdDev & 0,0024 & 0,0011 & 0,0014 & 31,758 & 88,566 & 88,2921 & 0,723 \\ \hline
        Median & 0,9986 & 0,9996 & 0,9998 & 5 & 2,2361 & 3,0811 & 0,723 \\ \hline
        25th quantile & 0,9977 & 0,9993 & 0,9995 & 2,8284 & 1,4142 & 1,7856 & 0,723 \\ \hline
       75th quantile & 0,9994 & 0,9999 & 0,9999 & 8,6023 & 3,5628 & 7,0533 & 0,723 \\ \hline

   \end{tabular}
   \vspace{2pt}
   \caption{\label{tab:f21.2} Final performance of FedCostWAvg in the 2021 FETS Challenge, Specificity, Hausdorff95 Distance and Communication Cost}
\end{adjustwidth}
\end{table}

\begin{table}[H]
\begin{adjustwidth}{0in}{0in}
    \centering
    \setlength\tabcolsep{3pt} 
    \begin{tabular}{|l|c|c|c|c|c|c|c|c|c|c|c|c|c|c|}
        \hline
        Label & Dice WT & Dice ET                & Dice TC         & Sens WT         & Sens ET         & Sens TC  \\ \hline 
        Mean & 0,76773526 & 0,741627265          & 0,769244434     & 0,749757737     & 0,770377324     & 0,765940502  \\ \hline  
        StdDev & 0,183035406 & 0,266310234       & 0,284212379     & 0,208271565     & 0,280923214     & 0,297081407 \\ \hline  
        Median & 0,826114563 & 0,848784494       & 0,896213442     & 0,819457864     & 0,886857246     & 0,893165349  \\ \hline   
        25quant & 0,700757354 & 0,700955694      & 0,739356651     & 0,637996476     & 0,728202272     & 0,73357786 \\ \hline   
        75quant & 0,897816734 & 0,910451814      & 0,943718628     & 0,905620122     & 0,956570051     & 0,964129538 \\ \hline  
        
    \end{tabular}
    \vspace{2pt}
    \caption{\label{tab:f22.1} Final performance of FedPIDAvg in the 2022 FETS Challenge, DICE and Sensitivity}
\end{adjustwidth}
\end{table}

\begin{table}[H]
\begin{adjustwidth}{0in}{0in}
    \centering
    \setlength\tabcolsep{3pt} 
    \begin{tabular}{|l|c|c|c|c|c|c|c|c|c|c|c|c|c|c|}
        \hline
        Label   & Spec WT    & Spec ET   & Spec TC           & H95 WT    & H95 ET          & H95 TC & Comm. Cost  \\ \hline
        Mean  & 0,9989230   & 0,9995742  & 0,999692          & 24,367549 & 32,796706    & 32,466108 & 0,300 \\ \hline
        StdDev  & 0,0016332  & 0,0007856   & 0,0006998       & 32,007897 & 89,31835     & 85,440174 & 0,300 \\ \hline
        Median   & 0,9994479   & 0,9997990  & 0,999868       & 11,57583  & 2,4494897     & 4,5825756 & 0,300 \\ \hline
        25th quant   & 0,998690  & 0,9995254  & 0,999695     & 5,4081799 & 1,4142135    & 2,2360679 & 0,300 \\ \hline
       75th quant  & 0,9998540   & 0,9999292   & 0,9999584   & 36,454261 & 10,805072    & 12,359194 & 0,300 \\ \hline

   \end{tabular}
   \vspace{2pt}
   \caption{\label{tab:f22.2} Final performance of FedPIDAvg in the 2022 FETS Challenge, Specificity, Hausdorff95 Distance and Communication Cost}
\end{adjustwidth}
\end{table}

    


    


    


\section{Conclusion}

This paper summarizes our winning contribution to the Federated Tumor Segmentation Challenge 2022. We submitted a PID-inspired aggregation strategy combined with a statistically inspired client selection. The aggregation function considers the number of training samples, the cost function decrease in the previous step as well as an integral term over the individual client's losses in the last rounds. The client selection models data center sizes as following a Poisson distribution and drops the outliers. Our method outperformed all other submissions.


%% file: acknowledgment.tex
\section*{\large{Acknowledgements}}
\noindent We appreciate the valuable input from our supervisors, David Naccache, Adrian Dalca, and Bjoern Menze. Moreover, we want to express our appreciation to the organizers of the Federated Tumor Segmentation Challenge 2022. Leon M\"achler is supported by the École normale supérieure in Paris. Johannes C. Paetzold is supported by the DCoMEX project, financed by the Federal Ministry of Education and Research of Germany. Suprosanna Shit and Ivan Ezhov are supported by the Translational Brain Imaging Training Network (TRABIT) under the European Union's `Horizon 2020' research \& innovation program (Grant agreement ID: 765148). With the support of the Technical University of Munich – Institute for Advanced Study, funded by the German Excellence Initiative. Ivan Ezhov is also supported by the International Graduate School of Science and Engineering (IGSSE). Johannes C. Paetzold and Suprosanna Shit are supported by the Graduate School of Bioengineering, Technical University of Munich.